\documentclass[10pt,twocolumn,letterpaper]{article}
\usepackage[pagenumbers]{cvpr} % To force page numbers, e.g. for an arXiv version

\usepackage{graphicx}
\usepackage{amsmath}
\usepackage{amssymb}
\usepackage{booktabs}
\usepackage{amsfonts}
\usepackage{bbding} %首先在导言区调用bbding包
\usepackage{color}
\usepackage{multirow}
\usepackage{multicol}
\usepackage[accsupp]{axessibility}
\usepackage{makecell}
\usepackage{colortbl} %xf
\usepackage{caption}
\definecolor{Gray}{gray}{0.92} %xf

\usepackage[pagebackref,breaklinks,colorlinks]{hyperref}

\usepackage[capitalize]{cleveref}
\crefname{section}{Sec.}{Secs.}
\Crefname{section}{Section}{Sections}
\Crefname{table}{Table}{Tables}
\crefname{table}{Tab.}{Tabs.}

\newcommand{\modelname}{DrivingDiffusion }

\def\cvprPaperID{0001} % *** Enter the CVPR Paper ID here
\def\confName{CVPR}
\def\confYear{2024}

\begin{document}

\title{DrivingDiffusion: Layout-Guided multi-view driving scene video generation with latent diffusion model}

\author{Xiaofan Li\textsuperscript{}, Yifu Zhang\textsuperscript{}, Xiaoqing Ye\textsuperscript{}\\
\textsuperscript{} Baidu Inc.\\
\small{Project page: \url{https://drivingdiffusion.github.io}}
} 

% \maketitle

\twocolumn[
\begin{center}  
\maketitle
\vspace{-2em}  
\centering  
\resizebox{1\linewidth}{!}{  
\includegraphics{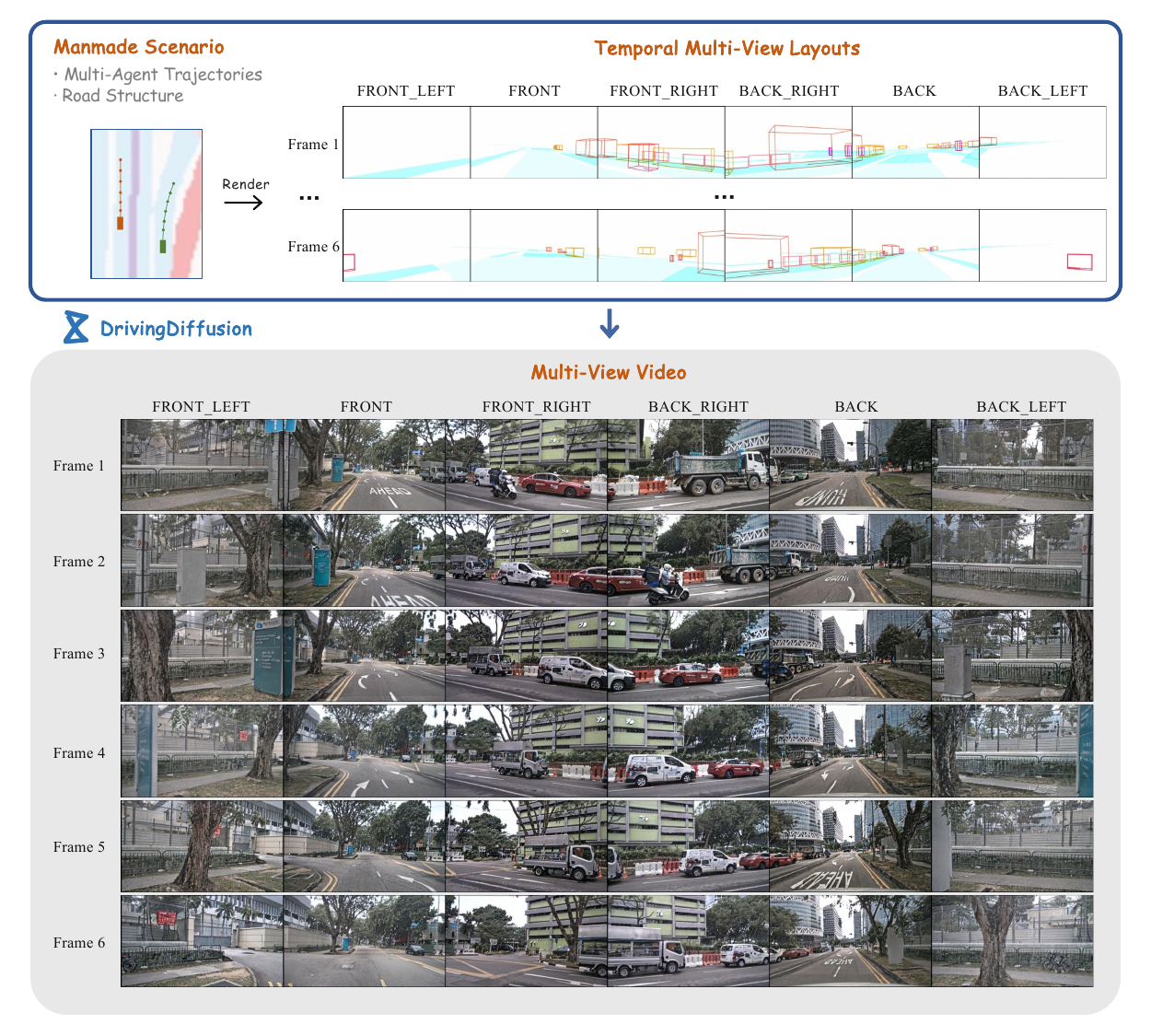}}  
\vspace{-2em}  
\captionof{figure}{Examples of the generated multi-view video frames from a man-made 3D layout (including the 3D bounding boxes of obstacles and the road structure) by DrivingDiffusion. The 3D layout is projected to the six camera views using camera parameters. We present the generated results of six camera views and six consecutive frames.}  
\label{fig:intro}
\end{center}]

%%%%%%%%% ABSTRACT
\begin{abstract}

With the increasing popularity of autonomous driving based on the powerful and unified bird's-eye-view (BEV) representation, a demand for high-quality and large-scale multi-view video data with accurate annotation is urgently required. However, such large-scale multi-view data is hard to obtain due to expensive collection and annotation costs. To alleviate the problem, we propose a spatial-temporal consistent diffusion framework DrivingDiffusion, to generate realistic multi-view videos controlled by 3D layout. There are three challenges when synthesizing multi-view videos given a 3D layout: How to keep 1) cross-view consistency and 2) cross-frame consistency? 3) How to guarantee the quality of the generated instances? Our \modelname solves the problem by cascading the multi-view single-frame image generation step, the single-view video generation step shared by multiple cameras, and post-processing that can handle long video generation. In the multi-view model, the consistency of multi-view images is ensured by information exchange between adjacent cameras. In the temporal model, we mainly query the information that needs attention in subsequent frame generation from the multi-view images of the first frame. We also introduce the local prompt to effectively improve the quality of generated instances. In post-processing, we further enhance the cross-view consistency of subsequent frames and extend the video length by employing temporal sliding window algorithm. Without any extra cost, our model can generate large-scale realistic multi-camera driving videos in complex urban scenes, fueling the downstream driving tasks. The code will be made publicly available.

\end{abstract}

%\vspace{-0.7cm}
\section{Introduction}

%------------------------FIG2---------
\begin{figure*}[t]
\centering
\includegraphics[width=1\linewidth]{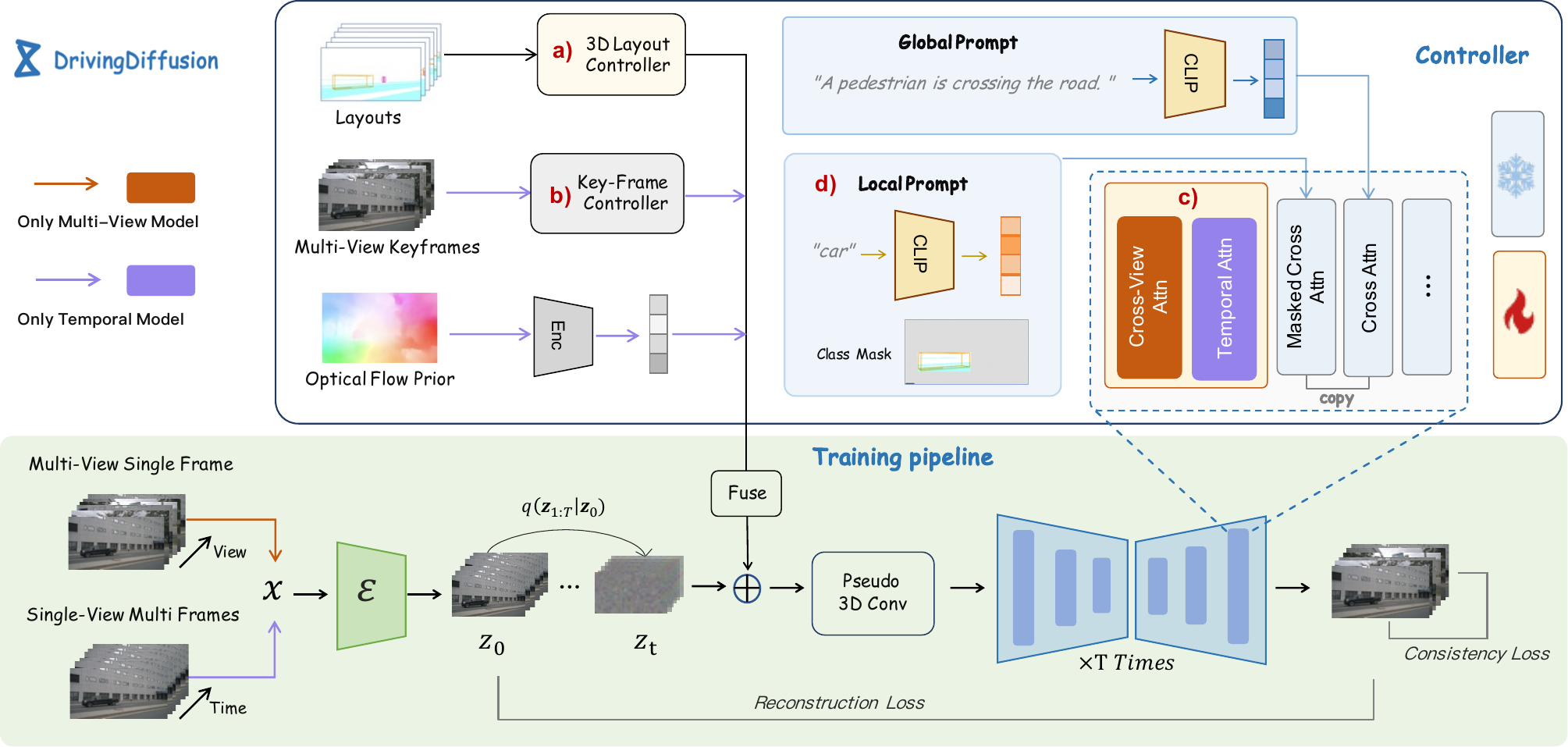}
\caption{Diagram of the multi-view video generation framework \modelname. For training, we separately train the multi-view model and the temporal model. These two models share similar structures, with the exception of the orange and purple components. During the inference stage, the two models are concatenated in a cascaded manner. First, the multi-view model generates the initial multi-view frame of the video. This frame is then set as the keyframe for the temporal model. Lastly, the temporal model generates video frames for each view, forming the final multi-view video.}
\centering
\label{fig:main}
\vspace{-10pt}
\end{figure*}
%------------------------FIG2---------

Autonomous driving has drawn extensive attention both from industry and academia for decades. The well-known bird's-eye-view (BEV) is a natural and straightforward candidate view to serve as a unified representation. BEV perception takes multi-view images as input and learns a unified BEV representation for downstream tasks such as 3D object detection \cite{li2022bevformer,liu2022petr,xiong2023cape}, segmentation \cite{lss,hu2021fiery,fang2023tbp}, tracking \cite{qd3dt,mutr3d,bytetrackv2} or trajectory prediction \cite{pnpnet,vip3d,pip}. High-quality and large-scale multi-view video data is critical for learning a powerful BEV representation for multiple perception tasks. However, high-quality data is often difficult to obtain due to the unaffordable labeling costs. As a result, there is a large demand for generating multi-view temporal-consistent video data that conforms to the real distribution and guarantees high consistency with 3D labeling and images.

We tackle the new task of multi-view video data generation in complex urban scenes. The generated data is controlled by a 3D layout around the ego-car. Figure \ref{fig:intro} shows our generated results from 3D layout to multi-view consecutive frames. Previous or concurrent works tackle similar tasks such as layout-to-image generation \cite{cheng2023layoutdiffuse,zheng2023layoutdiffusion}, single-view driving video generation \cite{blattmann2023align}. The most relevant task to ours is generating multi-view urban scene images from a BEV layout, \ie BEV segmentation mask, proposed in BEVGen \cite{swerdlow2023street}. Compared with these data generation tasks, our task is more complex (\ie multi-view vs. single-view, multi-frame vs. single-frame, complex urban scenes vs. simple road scenes), more controllable (\ie with 3D layout), and has wider applications. For example, it can serve as a kind of data augmentation for BEV detection models. By controlling the 3D layout, we can obtain a large amount multi-view videos for model training. Another application is editing specific objects in the video to simulate some rare scenarios such as collision. This helps the simulator to improve the safety of the driving system. 

There are three main challenges for the task of multi-view video data generation. First, we should maintain the cross-view consistency of the generated multi-view images of different cameras. For instance, the multi-view images in the same frame should appear like they were taken in the same physical location. Second, the cross-frame (or temporal) consistency should be maintained to produce a visually coherent video. The video frames should be temporally aligned. Third, it is significant to the guarantee quality of the generated instances (\ie vehicles and pedestrians) in the video. The instances are direct subjects of the perception tasks. 

We propose a multi-stage scheme \modelname to generate multi-view videos controlled by artificial 3D layout in complex urban scenes. It is based on the widely used image synthesis diffusion model. The 3D layout is utilized as additional control information for the diffusion model. \modelname contains a multi-view single-frame image generation model, a single-view temporal model shared by multiple cameras, and post-processing that includes a multi-view model specifically designed to enhance the consistency of subsequent generated frames and a sliding window processing that extends the video. In the multi-view model, we propose a cross-view attention module to ensure the consistency of multi-view images. An additional consistency loss is achieved by adding geometric constraints between different camera views. We also introduce local prompt to guide the relationship between the whole image and local instances, therefore effectively improving the quality of instance generation. In the single-view temporal model, we propose a Key-Frame controller that utilizes the first frame of the multi-view video sequence as a control condition to ensure cross-frame consistency. To avoid the need to train a separate video generation model for each camera, we embed statistics-based optical flow prior to each camera in the network. In post-processing, the "fine-tune" multi-view model we trained takes a subsequent generated frame and the first frame as conditions to further enhance the subsequent multi-view consistency. Finally, we use new keyframes in the sliding window to extend the video length.

In summary, our key contributions are as follows: 

(1) We address the new problem of multi-view video data generation from 3D layout in complex urban scenes. 

(2) We propose a generative model \modelname to ensure the cross-view, cross-frame consistency and the instance quality of the generated videos. 

(3) We achieve state-of-the-art video synthesis performance on nuScenes dataset. 

\section{Related Work}

\subsection{Text-to-Image Generation}
The text-to-image (T2I) generation task aims at generating realistic images based on text inputs. Early works solve the problem by turning it into a sequence-to-sequence problem. For example, DALL-E \cite{dalle} translates text tokens to discrete image embeddings obtained by VQ-VAE\cite{van2017neural}. The following works improve the quality of the generated images by more advanced architectures such as image tokenizers \cite{yu2022scaling}, encoder-decoder architectures \cite{yu2021vector} or hierarchical transformers \cite{ding2022cogview2}. Recently, Denoising Diffusion Probabilistic Models (DDPM) \cite{ho2020denoising} begin to conquer the T2I task. The quality of the generated image is further increased by utilizing the excellent ability to generate realistic images of diffusion models \cite{nichol2021glide,saharia2022photorealistic} or improving the text-image alignments \cite{ramesh2022hierarchical} using powerful text encoders \cite{clip}.

\subsection{Text-to-Video Generation}
The text-to-video (T2V) generation task generates video from the text descriptions Compared with the T2I task, the text-to-video (T2V) generation task is more challenging due to the temporally consistent characteristics involved. Some methods extend T2I methods to perform T2V by changing image tokens to video tokens \cite{wu2021godiva} or equipping the temporal attention modules to T2I models \cite{hong2022cogvideo}. The following work \cite{wu2022nuwa} proposes an auto-regressive framework to solve both T2I and T2V tasks. More recently, Video Diffusion Models \cite{ho2022video} utilizes a space-time factorized U-Net with joint image and video data training. Make-A-Video \cite{singer2022make} directly translates the tremendous recent progress in T2I to T2V by introducing an effective spatial-temporal module on top of T2I models. Similar to Make-A-Video, we also adopt pre-trained T2I models to generate multi-view videos. 

\subsection{Layout-to-Image generation}
This task can be seen as the reverse process of object detection or segmentation, which generates images with the input of bounding boxes or segmentation maps. Previous works utilize GANs \cite{he2021context,li2021image,park2019semantic,sylvain2021object} or diffusion models \cite{cheng2023layoutdiffuse,zheng2023layoutdiffusion} to generate synthetic images based on 2D layout. These methods encode the layout as a condition image that is downsampled and upsampled jointly with the data. Recently, BEVGen \cite{swerdlow2023street} generates multi-view urban scene images from BEV layout based on VQ-VAE \cite{van2017neural}. Compared with BEVGen, we tackle a more complex task that generates temporally consistent multi-view videos. We also adopt diffusion models \cite{ho2022video} to generate more realistic videos.

\section{Method}
In this section, we begin with a brief description of the denoising diffusion probabilistic model (DDPM) and the latent diffusion model (LDM) on which our method is based, and this can be found in Section \ref{sec:preliminary}. Subsequently, we introduced \modelname's pipeline: the cross-view model, temporal model and post-processing, including training and inference strategy, which is discussed in Section \ref{sec:train}. We also present our methods for enhancing cross-view and cross-frame consistency in the video as well as our local prompt design for improving instance generation quality in Section \ref{sec:consistency} and Section \ref{sec:lp}, respectively. An overview of our approach is shown in Figure \ref{fig:main}.

\subsection{Preliminary: Latent Diffusion Models}
\label{sec:preliminary}
Latent diffusion models (LDM)~\cite{rombach2022high} are a type of diffusion model that models the distribution of the latent space of images and have recently shown remarkable performance in image synthesis. The LDM consists of two models: an autoencoder and a diffusion model. 

The autoencoder learns to compress and reconstruct images using an encoder $\mathcal{E}$ and a decoder $\mathcal{D}$. 

The encoder first projects the image $x$ to a lower-dimensional latent space $z$, and the decoder then reconstructs the original image from the latent space, resulting in $\tilde{x} = \mathcal{D}(z)$.  

Then the latent generative model is trained to recreate a fixed forward Markov chain $x_1, \ldots, x_T$ via DDPMs~\cite{ho2020denoising}.

Given the data distribution $z_0 \sim q(z_0)$, the Markov transition $q(z_t|z_{t-1})$ is defined as a Gaussian distribution with a variance schedule $\beta_t \in (0, 1)$, which can be formulaic as,
\begin{equation}
% \begin{aligned}
    q(z_t|z_{t-1}) = \mathcal{N}(z_t; \sqrt{1-\beta_t} z_{t-1}, \beta_t \mathbb{I}), \quad t = 1, \ldots, T.
% \end{aligned}
\end{equation}

According to the Bayes' rule and the Markov property, we can derive explicit expressions for the conditional probabilities $q(z_t|z_0)$ and $q(z_{t-1}|z_t, z_0)$ as follows.

% {\small
% \begin{equation}

%%origin-start
% \begin{align}
%     q(z_t|z_0) & = \mathcal{N}(z_t; \sqrt{\Bar{\alpha}_t} z_0, (1 - \Bar{\alpha}_t) \mathbb{I}), \quad t = 1, \ldots, T, \\
%     q(z_{t-1}|z_t, z_0) & = \mathcal{N}(z_{t-1}; \Tilde{\mu}_t(z_t, z_0), \Tilde{\beta}_t \mathbb{I}), \quad t = 1, \ldots, T, \\
%     w.r.t.~~~~ &\alpha_t = 1 - \beta_t, ~ \Bar{\alpha}_t = \prod_{s=1}^t \alpha_s, ~ \Tilde{\beta}_t = \frac{1 - \Bar{\alpha}_{t-1}}{1 - \Bar{\alpha}_t} \beta_t, \\
%     &\Tilde{\mu}_t(z_t, z_0) = \frac{\sqrt{\Bar{\alpha}_t}\beta_t}{1 - \Bar{\alpha}_t} z_0 + \frac{\sqrt{{\alpha}_t} (1 - \Bar{\alpha}_{t-1})}{1 - \Bar{\alpha}_t} z_t.
% \end{align}
%%origin-end

%new
\begin{align}
    q(z_t|z_0) &= \mathcal{N}(z_t; \sqrt{\Bar{\alpha}_t} z_0, (1 - \Bar{\alpha}_t) \mathbb{I}), \quad t = 1, \ldots, T, \\
    q(z_{t-1}|z_t, z_0) &= \mathcal{N}(z_{t-1}; \Tilde{\mu}_t(z_t, z_0), \Tilde{\beta}_t \mathbb{I}), \quad t = 1, \ldots, T, \\
    w.r.t.~~~~ \alpha_t = & 1 - \beta_t, ~ \Bar{\alpha}_t = \prod_{s=1}^t \alpha_s, ~ \Tilde{\beta}_t = \frac{1 - \Bar{\alpha}_{t-1}}{1 - \Bar{\alpha}_t} \beta_t, \\
    \Tilde{\mu}_t(z_t, z_0) & = \frac{\sqrt{\Bar{\alpha}_t}\beta_t}{1 - \Bar{\alpha}_t} z_0 + \frac{\sqrt{{\alpha}_t} (1 - \Bar{\alpha}_{t-1})}{1 - \Bar{\alpha}_t} z_t.
\end{align}

% \end{equation}
% }%
DDPMs leverage the reverse process with a prior distribution $p(z_T) = \mathcal{N}(z_T; 0, \mathbb{I})$ and Gaussian transitions to generate the Markov chain $z_1, \ldots, z_T$, 

{\small
\begin{equation}
    p_\theta(z_{t-1}|z_t) = \mathcal{N}(z_{t-1}; \mu_\theta (z_t, t), \Sigma_\theta(z_t, t)), \quad t = T, \ldots, 1.
\end{equation}}%

Here $\theta$ is the learnable parameter we use to ensure that the generated reverse process is close to the forward process. 

DDPMs follow the variational inference principle by maximizing the variational lower bound of the negative log-likelihood, which has a closed form given the KL divergence among Gaussian distributions. Empirically, these models can be interpreted as a sequence of weight-sharing denoising autoencoders $\epsilon_\theta(z_t, t)$, which are trained to predict a denoised variant of their input $z_t$. The objective can be simplified as follows.

% {\small
\begin{equation}
     \mathbb{E}_{z, \epsilon \sim \mathcal{N}(0, 1), t} \left[ \| \epsilon - \epsilon_\theta(z_t, t) \|^2_2 \right]. 
\end{equation}
%}%
\vspace{-1em}

\subsection{\modelname}
\label{sec:train}
%---
\paragraph{Overview.}
% image diffusion
The prevalent diffusion model for image synthesis typically utilizes a U-Net architecture for denoising, which entails two rounds of spatial downsampling followed by upsampling. It comprises numerous layers of 2D convolution residuals and attention blocks, containing self-attention, cross-attention, and a feedforward network. Spatial self-attention evaluates the interdependence between pixels within the feature map, whereas cross-attention considers the alignment between pixels and conditional inputs, such as text.
% video diffusion
When generating videos, it is customary to utilize an elongated $1\times3\times3$ convolution kernel for the video input to retain the parameters of the single-frame pre-trained diffusion model and expand them to novel dimensions. In our multi-view and temporal model, we introduce additional dimensions for view and time separately. Similar to VDM \cite{ho2022video}, we expand the 2D convolutional layer into a pseudo-3D convolutional layer with a $1\times3\times3$ kernel and integrate a time self-focusing layer in each transformer block to facilitate inter-dimensional information exchange. Furthermore, we have incorporated several modules, such as the 3D layout controller, key-frame controller, cross-view / cross-frame consistency module, and a local prompt guidance module, to enhance instance generation quality. Figure \ref{fig:main} presents an overview of our proposed DrivingDiffusion. Our experimental findings indicate that this technique can generate multi-view videos that are accurately aligned with 3D layout while exhibiting data quality comparable to real-world data.

\paragraph{3D Layout Controller.}
In order to ensure an accurate mapping between the target instance position in the physical space around the ego-car and the generated image, we project the road structure and each instance onto the 3D layout image using the parameters of the camera. It is important to mention that we incorporate the road-structure information, the target category, and the target instance ID corresponding to each pixel, and encode this information into RGB values. Next, the aforementioned input is fed into a ResNet-like 3D layout controller to encode the U-Net model at different resolutions (\ie 64 × 64, 32 × 32, 16 × 16, 8 × 8) corresponding to different levels of features. We inject this additional control information at different resolutions into each layer of the U-Net model through residual connections, as is shown in a) Figure~\ref{fig:main}. The advantage of this type of control mechanism is that it is able to influence the diffusion model from various receptive fields without introducing too many parameters (i.e., a trainable copy of ControlNet \cite{zhang2023adding}).

\paragraph{Multi-View Model. } 
For the multi-view single-frame model, we employed the 3D layout of all views as input, along with textual descriptions of the scene, to generate highly aligned, multi-view images. At this stage, we only utilized one 3D layout controller. 

\paragraph{Temporal Model. }
As for the temporal single-perspective model, we input the 3D layout of all video frames from a single perspective, the first frame image of the video sequence, and the optical flow prior corresponding to the camera. The output is a temporally consistent video from a single viewpoint. Specifically, the added dimension in our model changes from multiple views to multiple frames. Additionally, we include a Key-Frame controller that utilizes the first frame of the multi-view video sequence as a control condition for generating consistent videos. To avoid the need to train a separate video generation model for each camera, statistical optical flow trend at each camera view, with a fixed value as a prior and encoded as a token, is input into the model in the same way as time embedding.

%FIG-pipe start
%[Option] fine-tune model &latent frame interpolation model.
\begin{figure}
    \centering
    \includegraphics[width=1\linewidth]{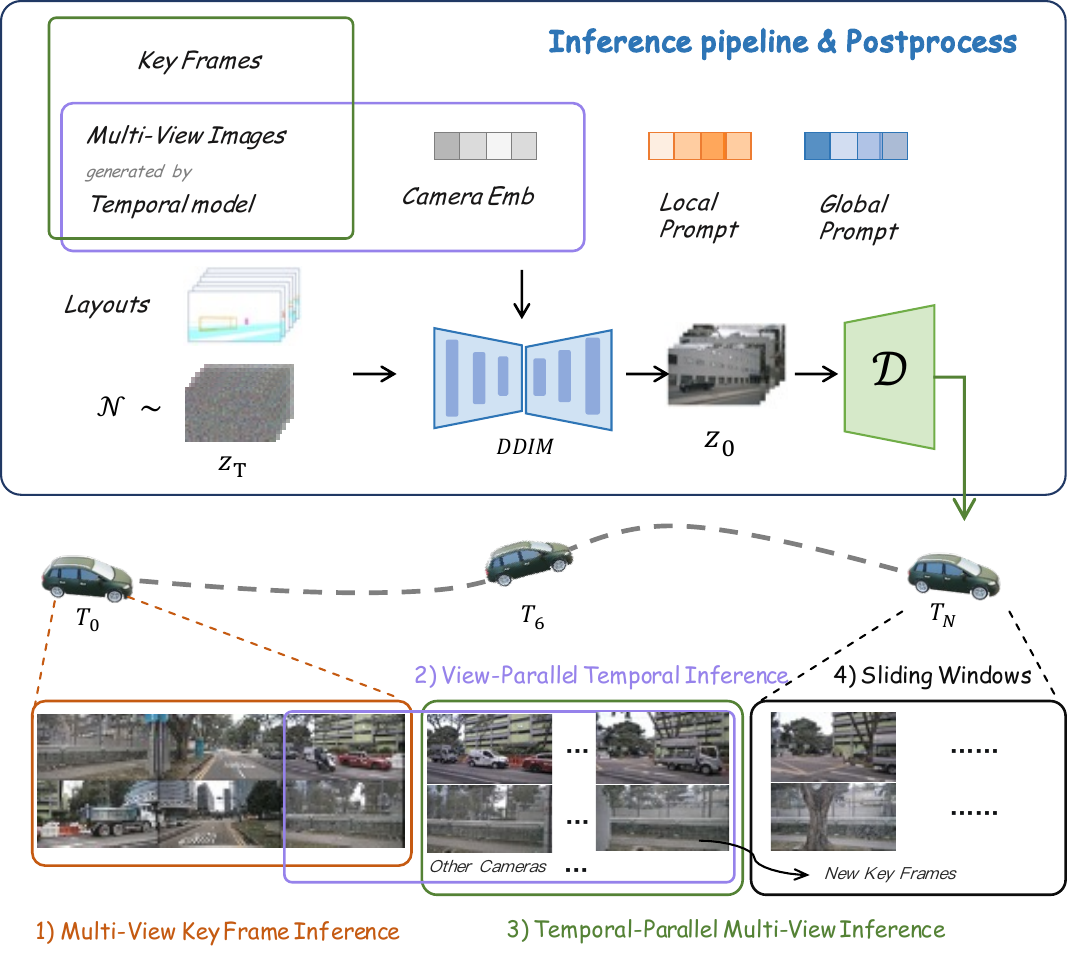}
    \caption{Multi-view Long video generation pipeline.}
    \label{fig:pipe}
\end{figure}
%FIG-pipe end

\paragraph{Post-Processing. }
In our temporal model, we take the multi-view images at the initial moment as the keyframe. This is because, in a short period of time, we assume that most of the information needed for subsequent frames of each view can be found in the multi-view images at the initial moment. However, when the time interval is too long, there is still a probability that multiple views of the subsequent frames of the video may be inconsistent due to factors such as occlusion. To handle long sequence videos, we train a finetune model after the cross-view and temporal models to enhance the consistency of the following frames. Concretely, the finetune model utilizes the existing 
 structure and parameters of the cross-view model, with the exception of an extra controller for inputting frames generated by the temporal model. In the training, the finetune model uses a large number of temporal subsequent frames generated by the temporal model through the original multi-view images in the dataset as inputs, and the corresponding original images in the dataset as truth values. During inference, the model inputs the multi-view images of subsequent frames of the generated video and outputs the images after finetuning. Since it is only engineering work to improve the generation quality, we see it as a part of post-processing for long video generation tasks. 

Finally, we take a certain moment of subsequent multi-view images as new keyframes and conduct temporal inference again. The video length is extended with the idea of sliding window.

\paragraph{Multi-stage pipeline for long video. }
We introduce a multi-stage inference strategy to generate multi-view long videos, as shown in Figure \ref{fig:pipe}: 1) We first adopt the multi-view model to generate the first frame panoramic image of the video sequence. 2) Then we use the generated image from each perspective as input for the temporal model, allowing for parallel sequence generation for each corresponding viewpoint. 3) For subsequent frames, we employ the finetune model for parallel inference as well. 4) Extend the video after identifying new keyframes, just like the sliding window algorithm does. Finally we obtain the entire synthetic multi-view video.

%------------------------3.3---------
\subsection{Cross-View and Cross-Frame Consistency}
\label{sec:consistency}
As described in section 3.2, the input of 3D layout provides the model with the location, category, and instance information of each target, to some extent, aiding in enhancing local multi-view and temporal consistency. However, this alone is insufficient to ensure global consistency, particularly in static scenes. To address this, we propose an efficient attention mechanism that emphasizes cross-view and cross-frame interactions and introduces geometric constraints to supervise generated results.

\begin{figure}
    \centering
    \includegraphics[width=1\linewidth]{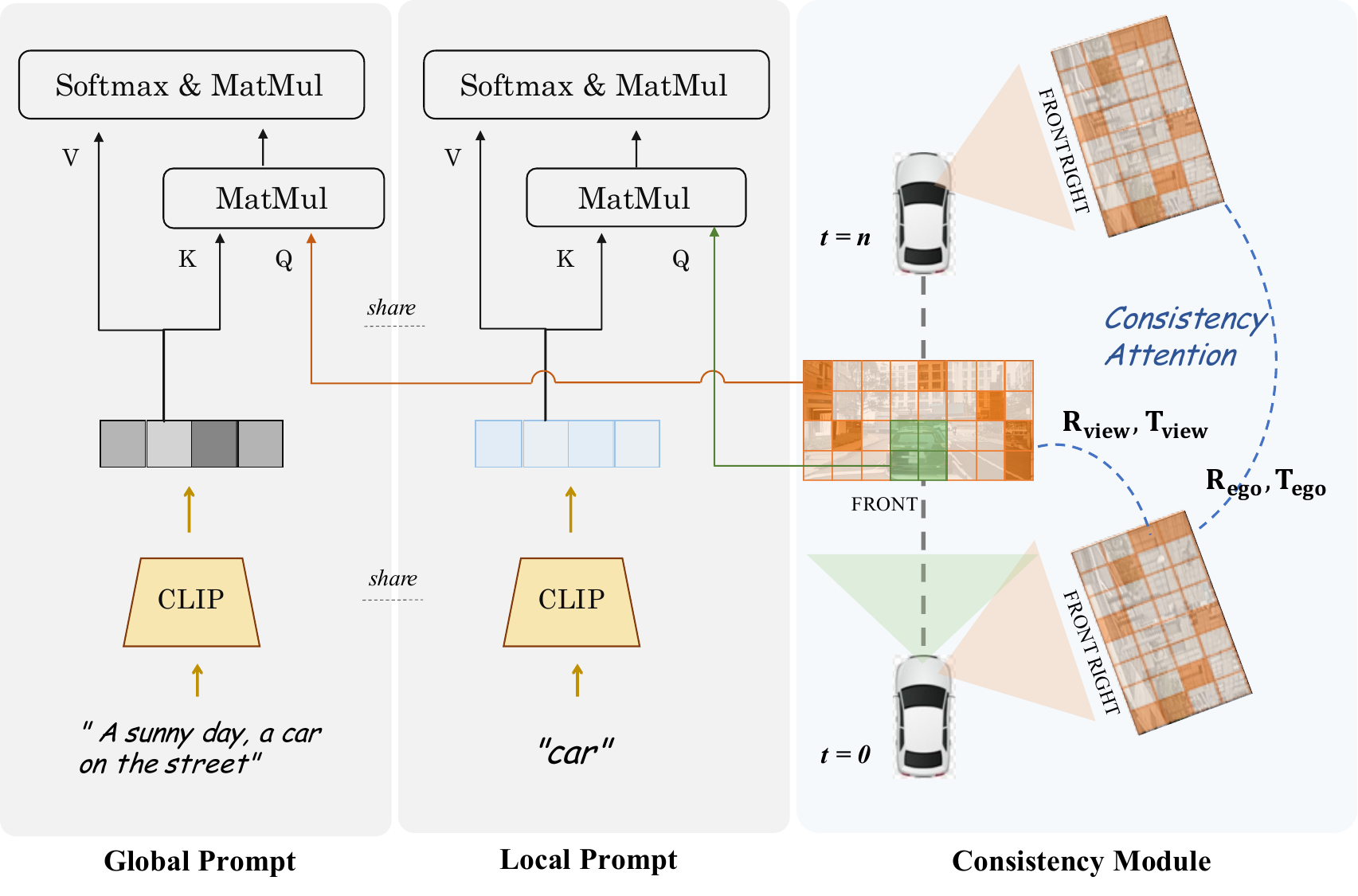}
    \caption{A schematic of the Consistency module(the right part) and local prompt(the left part) methods.}
    \label{fig:detail}
\end{figure}

The right part of Figure \ref{fig:detail} shows the range in which the Consistency Module is in effect: (1) adjacent views in multi-view, (2) the current frame and the first/previous frames in multi-frame.

%FIG-detail start
%[Option] fine-tune model &latent frame interpolation model.
%FIG-detail end

\paragraph{Consistency Attention. } 
%%% todo：公式
% Figure \ref{fig:main} shows the consistency module. 
For the cross-view model, we posit that each perspective should primarily focus on its adjacent left and right perspectives. The latent representation of a view and its left and right neighbors in the multi-view model $v$ are denoted as $z_{v}^{i}$, $z_{v}^{i-1}$, and $z_{v}^{i+1}$.

We attach a consistency attention layer to each attention block to model the new dimensions. For the multi-view model, in order to enhance the cross-view consistency, cross-view attention was calculated using the latent expression of the \textit{i th} view $z_{v}^{i}$ and the sum of the latent expression of the two adjacent views $z_{v}^{i-1}$ and $z_{v}^{i+1}$. Our cross-view consistency attention is formulated as:

\begin{small}
\begin{equation}
\begin{aligned}
&  Attention_{cross-view}(Q_v,K_v,V_v) = Softmax(
\\ &\frac{\left(W^{Q_v} z_v^i \right) \left( W^{K_v} \left[ z_{v}^{i-1}, z_{v}^{i+1} \right] \right)^T}
{\sqrt{d}}) \cdot \left(W^{V_v} \left[ z_{v}^{i-1}, z_{v}^{i+1} \right]\right)
\end{aligned}
\end{equation}
\end{small}

% }
where $\left[ , \right]$ denotes concatenation operation. $W^{Q_v}$, $W^{K_v}$, and $W^{V_v}$ are learnable matrices in the cross-view model that project the inputs to query, key and value, respectively, and $d$ is the output dimension of key and query features. 

For the temporal model, the \textit{j$th$} frame in the temporal model $t$ such that $j > 1$ is denoted as $z_{t}^{j}$. Cross-frame attention was calculated using the sum of the latent representation of the \textit{j$th$} frame and the sum of the first frame with the \textit{(j-1)$th$} frame. Similarly, we implement cross-frame consistency attention as:

\begin{small}
\begin{equation}
\begin{aligned}
& Attention_{cross-frame}(Q_t,K_t,V_t) = Softmax(
\\ &\frac{\left(W^{Q_t} z_t^j \right) \left(W^{K_t} \left[ z_{t}^{1}, z_{t}^{j-1} \right] \right)^T}
{\sqrt{d}}) \cdot \left(W^{V_t} \left[ z_{t}^{1}, z_{t}^{j-1} \right]\right)
\end{aligned}
\end{equation}
\end{small}

where $W^{Q_t}$, $W^{K_t}$, and $W^{V_t}$ are learnable matrices in the temporal model.

\paragraph{Consistency Loss. }
We supervise the consistency of the generated results using a pre-trained network $\mathcal{F}$ that can perform image matching and regression relative poses. During training, we fix the parameter of $\mathcal{F}$ and use the actual relative pose as the ground truth. For the cross-view model, we supervise the pose between adjacent frames, and for the temporal model, we supervise the adjacent frames in a time sequence.

%------------------------FIG---------
\begin{figure*}[t]
\centering
\includegraphics[width=0.99\linewidth]{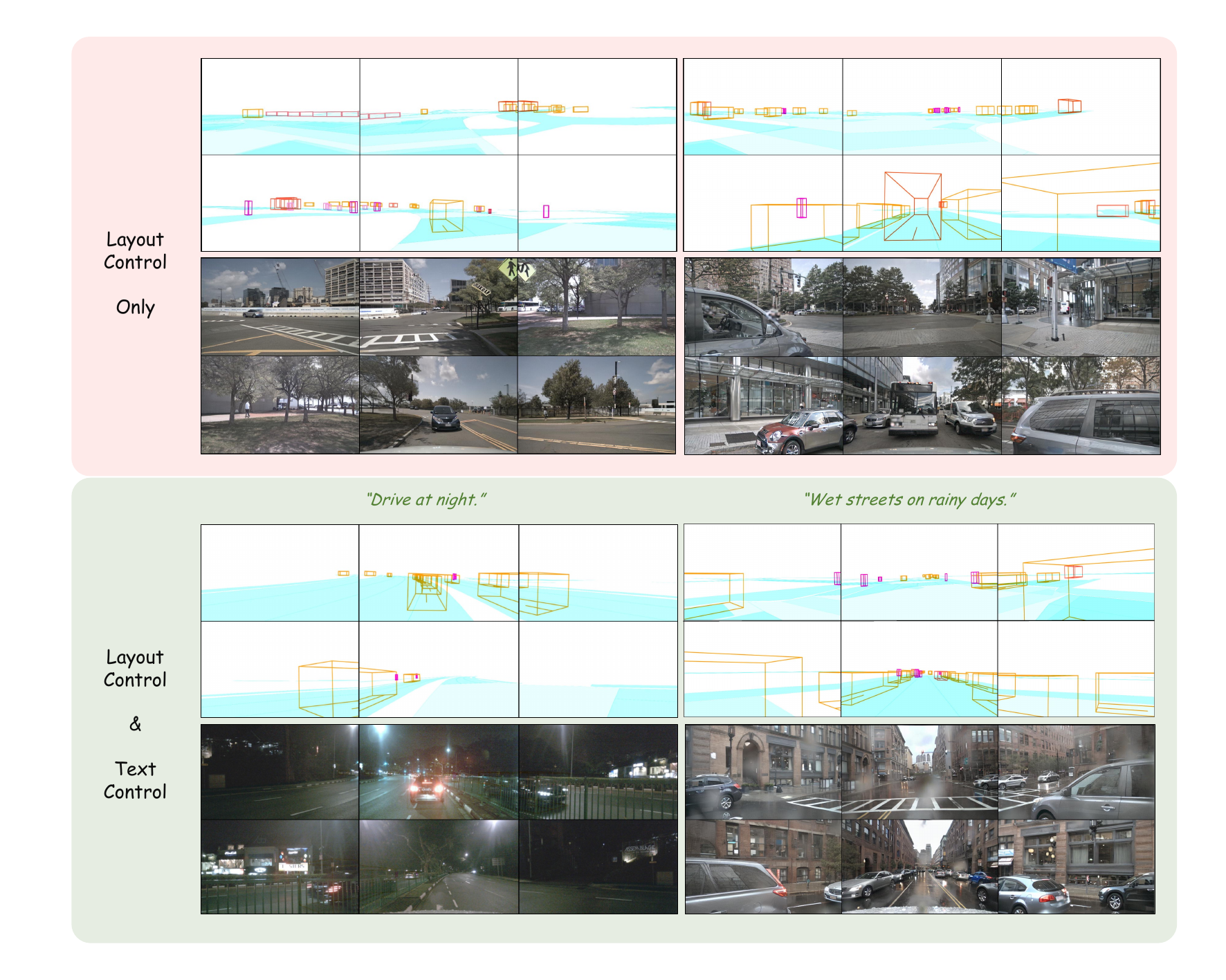}
\vspace{-10pt}
\caption{Street view generation results of the multi-view model. We adopt the 3D layout to generate the images of 6 surrounding cameras in 4 different scenarios. Results with and without text control are displayed separately. }
\centering
\label{fig:show1}
% \vspace{-10pt}
\end{figure*}
%------------------------FIG---------

%------------------------FIG temporal---------
\begin{figure*}[t]
\centering
\includegraphics[width=0.99\linewidth]{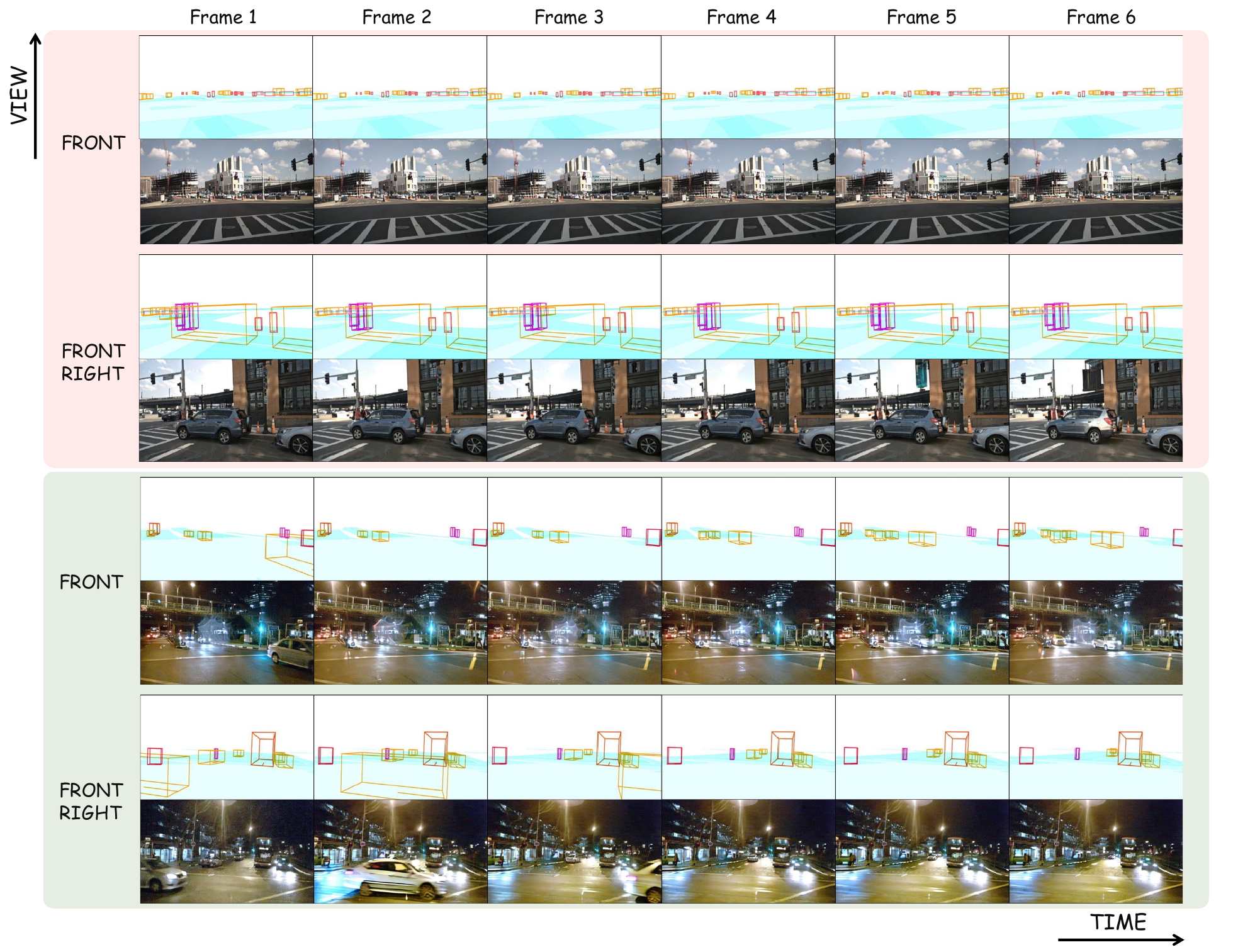}
\caption{Street view generation results of the temporal model. We show the visualization results of 6 consecutive frames of 2 adjacent views. }
\centering
\label{fig:show2}
% \vspace{-10pt}
\end{figure*}
%------------------------FIG---------

%------------------------3.4---------
\subsection{Local Prompt}
\label{sec:lp}

To enhance the generation quality of instances, we propose the local prompt module. We can refer to d) of Figure \ref{fig:main}. First, we store the category $k$ and the text $T_k$ corresponding to the category name in advance, and encode the category embedding $F_k = \phi(T_k)$, where $\phi$ is the CLIP \cite{clip} encoder. We utilize the smallest surrounding rectangular area of the projected 3D layouts as the mask $M_k$ for each category $k$. Cross attention is then computed using latent representation computation $z$ and category text encoding $E_k$ in the same way as the global prompt, using $M_k$ as the mask for attention. The left part of Figure \ref{fig:detail} illustrates how global prompt and local prompts cooperate with each other.

We did not employ fixed digit encoding during training, as we hypothesized that the concepts of local and global prompts were well-aligned, with the only difference being their respective scopes. Therefore, our local prompt replicates the same structure and parameters as the global prompt, enabling it to fully leverage the parameters of the pre-trained diffusion model, which already encompasses an understanding of our target class.

\section{Experiment}
\label{sec:exp}
In this section, we will evaluate \modelname based on several metrics. Firstly, we will provide a detailed description of the experimental setup, dataset, and evaluation criteria. Secondly, we will analyze our method qualitatively and compare it quantitatively with other methods
. Lastly, we will conduct an ablative analysis, which will be thoroughly discussed and analyzed.

%------------------------4.1---------

\subsection{Experiment Details}
\label{sec:Dataset}

\paragraph{Dataset}
We present the results of generated images from all six camera views trained on the nuScenes dataset. We compared the image quality of our method with other multi-view single-frame approaches, and the video quality with other single-view video methods. To evaluate our work, we utilized metrics such as FID and FVD, along with quantitative demonstrations highlighting the proposed method's advantages for autonomous driving. Furthermore, we qualitatively showcased our work through visualization. In multi-view visualizations, we horizontally flipped the back left and right cameras to emphasize our model's consistency across images. Consequently, the side, front, and back cameras align at their outer edges in all figures.

\paragraph{Details}
We implement our approach based on the official code base for stable diffusion \cite{rombach2022high} and the publicly available 1.4 billion parameter T2I model. We freeze the image autoencoder and encode each view or video frame individually as a potential representation. We trained the cross-view model on nuScenes for 50,000 steps, the temporal model for 40,000 steps, and the fine-tune model for 10,000 steps. Six groups of six cameras with a resolution of 512 × 512 were selected for training. The training process was carried out on eight NVIDIA Tesla 40G-A100 GPUs, and both the cross-view model and cross-frame model could be completed within 36 hours. In terms of reasoning, we used a DDIM sampler \cite{song2020denoising} to generate a multi-view video guided by 3D layouts in the experiment.

%------------------------4.2---------
\subsection{Experimental Results}
\label{sec:Results}

%-------------------------
\paragraph{Qualitative Results}

We demonstrate the quality and the consistency of single-frame multi-view and single-view video generation in Figure \ref{fig:show1} and Figure \ref{fig:show2}, respectively. We can see that our model obtains both cross-view and cross-frame consistency. The generated instances are accurately controlled by the input 3D layout. When the 3D layout of the vehicle category is very close to the main vehicle, the layout tends to be projected into the images of two perspectives, and even some images are out of the scope. In this case, our model can still generate cross-view consistent adjacent vehicles. For the quality of instance generation, whether it is the non-rigid pedestrian class, highly occluded/truncated vehicles, or the cone bucket class with less sample size in the dataset, our model can give high-quality instance generation results and maintain self-consistency with the surrounding environment.

%%%%%%%%%%%%%
\begin{table}[!t]
% \small
\setlength{\tabcolsep}{3mm}
\resizebox{\linewidth}{!}{
\begin{tabular}{l|c c| c c} 
 \hline
     Method    & Multi-View & Multi-Frame & FID$\downarrow$    & FVD$\downarrow$ \\ %[0.5ex] 
      \hline
     BEVGen~\cite{lang2019pointpillars}  &   \Checkmark &  & 25.54  & - \\
     DriveDreamer~\cite{2309.09777}  &   & \Checkmark & 52.6  & 452 \\
     DrivingDiffusion  & \Checkmark &  &15.89  &  - \\
     DrivingDiffusion  & &\Checkmark & 15.85 &  335 \\
     \rowcolor{Gray} DrivingDiffusion  & \Checkmark & \Checkmark & \textbf{15.83}  &  \textbf{332} \\
 \hline
\end{tabular}
}
\caption{Comparison with other methods on nuScenes validation. The FID indicator and FVD indicator feed back the image and video quality, respectively.}
\label{tab:nuScenes-test} % end caption
\vspace{-0.2cm}
\end{table}

%%%%%%%%%
\definecolor{codegreen}{rgb}{0.0,0.6,0.0}
\begin{table}
  \centering
  \resizebox{0.48\textwidth}{!}{
      \begin{tabular}{l|cc|cc}
        \toprule
        Method & Road mIoU$\uparrow$ & Vehicle mIoU$\uparrow$  & Drivable mIoU$\uparrow$  & Object NDS$\uparrow$\\     \midrule
                Baseline& 71.3 & 36.0 & 81.7 & 41.2 \\     
                BEVGen & 50.2 \textcolor{codegreen}{(-21.1\%)}& 5.9 \textcolor{codegreen}{(-39.1\%) }& - & - \\     
                Sparse BEVGen & 50.9 \textcolor{codegreen}{(-20.4\%)}& 6.7 \textcolor{codegreen}{(-29.4\%)}& - & -\\
                \rowcolor{Gray}\modelname & \textbf{63.2}\textcolor{codegreen}{(-8.1\%)}& \textbf{31.6}\textcolor{codegreen}{(-4.4\%)}& \textbf{67.8} \textcolor{codegreen}{(-13.9\%)}& \textbf{33.1} \textcolor{codegreen}{(-8.1\%)}\\
        \bottomrule
        \end{tabular}}
  \caption{Detection and segmentation results of different validation data over all 6 views on the validation set of nuScenes. In green is the relative drop compared with the standard nuScenes validation data. Road mIoU and Vehicle mIoU are obtained by the CVT \cite{zhou2022cross} segmentation model while Drivable mIoU and Vehicle NDS are obtained by the BEVFusion \cite{liu2022bevfusion} multi-task model.}
    \label{tab:bevfusion} % end caption
    \vspace{-1.5em}
\end{table}

\vspace{-1.5em}

\paragraph{Quantitative Results} 

To assess the performance of our models, we employ the frame-wise Fréchet Inception Distance (FID) \cite{parmar2021cleanfid} to evaluate the quality of generated images. For video quality evaluation, we utilize the Fréchet Video Distance (FVD) \cite{unterthiner2018towards}. Unless otherwise noted, all metrics are calculated on the nuScenes validation set. 

Due to the lack of related work, we have sorted out the recent work and compared the quality of image/video generation under various tasks. For example, tasks of image generation in autonomous driving scenes like BEVGen \cite{swerdlow2023street} and tasks of video generation like DriveDreamer \cite{2309.09777}.

As presented in Table~\ref{tab:nuScenes-test}, we made every effort to use the most closely related metrics and settings for comparison. Despite generating both multi-view and multi-frame outputs without performing frame interpolation or super-resolution, our method achieved FID and FVD scores of 15.83 and 332, respectively. These results demonstrate considerable advantages over other approaches. The last three rows in Table~\ref{tab:nuScenes-test} show that the indexes of the complete DrivingDiffusion and multi-view/temporal model are similar, proving our method's stability.

Although methods such as FID are commonly used to measure the quality of image synthesis, they do not fully capture the design objectives of our task, nor do they reflect the quality of synthesis across different semantic categories. Since we are trying to generate multi-view images consistent with the 3D layout, we want to measure our performance on this consistency. 

We take advantage of the official model provided by CVT \cite{zhou2022cross} and BEVFusion\cite{liu2022bevfusion} as evaluators. We adopted the same set of generated images conditioned on a ground-truth 3D layout as the nuScenes validation set, applied CVT and BevFusion to each set of generated images, and then compared the predicted layout to the ground-truth BEV layout. We report average intersection crossing (mIoU) scores for drivable areas and NDS for all the object classes in Table~\ref{tab:bevfusion}.

%%%%%%%%%%%%% BEVFUSION-AUG
\begin{table}
\vspace{0.5cm}

% \small
\setlength{\tabcolsep}{3mm}
\centering
\resizebox{0.9\linewidth}{!}{
\begin{tabular}{l|c | c c} 
 \hline
     Setting& Extra-Data Amount & Object NDS$\uparrow$ & mAOE$\downarrow$   \\ %[0.5ex] 
      \hline
     (a)& 0 & 0.412 & 0.5613\\
     (b)& 2,000 & 0.418 & \textbf{0.5295} \textcolor{codegreen}{(-3.18\%)} \\
     \rowcolor{Gray}(c)&  6,000 & \textbf{0.434} \textcolor{codegreen}{(+2.2\%)} & \textbf{0.5130} \textcolor{codegreen}{(-4.83\%)}\\
 \hline
\end{tabular}
}
\caption{The impact of data augmentation using synthetic data on perceptual performance. We use the BEVFusion visual model as the baseline.}
\label{tab:aug} % end caption
\vspace{-0.2cm}
\end{table}

%%% ablation
% \vspace{-1.5em}
\begin{table}
% \small
\setlength{\tabcolsep}{3mm}
\centering
\resizebox{\linewidth}{!}{
\begin{tabular}{l|c c| c c c} 
 \hline
     Setting & Consistency Module & Local Prompt & FID$\downarrow$    & FVD$\downarrow$ & Object NDS$\uparrow$\\ %[0.5ex] 
      \hline
     (a)&\ & \ & 17.30& 420 &28.1\\
     (b)&\Checkmark &  & 16.67& 365&28.9\\
     (c)&\ & \Checkmark & 17.62& 409&32.6\\
     \rowcolor{Gray}(d)&\Checkmark & \Checkmark & \textbf{15.83}  &  \textbf{332} & \textbf{33.1}\\
 \hline
\end{tabular}
}
\caption{Ablation of the consistency module and the local prompt.}
\label{tab:ab1} % end caption
\vspace{-0.2cm}
\end{table}

To ensure a fair comparison, we focus on the drop compared to the standard nuScenes validation data. The green numbers in Table \ref{tab:bevfusion} represent this drop. Observing the road generation quality, we can see that \modelname significantly outperforms BEVGen, with a difference of -8.1\% compared to BEVGen's -21.1\%. Similarly, for the instance generation quality, \modelname surpasses BEVGen by a substantial margin, with a difference of -4.4\% compared to BEVGen's -39.1\%. These remarkable improvements demonstrate the effectiveness of our proposed approach.

%add1  - - aug
Table \ref{tab:aug} presents the performance improvement achieved in the BEV perception task through synthesized data-augmentation. In the original training data, there was an issue with long-tail distribution, specifically with small targets, close-range vehicles, and vehicle orientation angles. We focused on generating additional data for these categories with limited samples to address this.
%ab
In Setting (b), we only added 2,000 frames of synthetic data, focusing on improving the distribution of obstacle orientation angles. Comparing Setting (a) and (b), and using BEVFusion as the baseline, although NDS only showed a slight improvement, mAOE significantly decreased from 0.5613 to 0.5295. 
%ac
Comparing Setting (a) and (c), by augmenting the data with 6,000 frames of synthetic data, which is more focused on rare scenarios, we observed notable enhancements on the nuScenes validation set. The NDS results increased from 0.412 to 0.434, while the mAOE decreased from 0.5613 to 0.5130. This proves that the data augmentation of synthetic data brings significant benefits to perceptual tasks.

\vspace{0.4cm}

\subsection{Ablation Study}
\label{sec:ablation}

To verify the validity of our design choices, an ablation study was performed on the key features of the model, including the consistency module and the local prompt. We run these experiments on the same nuScenes validation set in Table \ref{tab:ab1}. % and Table \ref{tab:ab2}. 

%------------------------FIG control---------
\begin{figure*}
\centering
\includegraphics[width=0.99\linewidth]{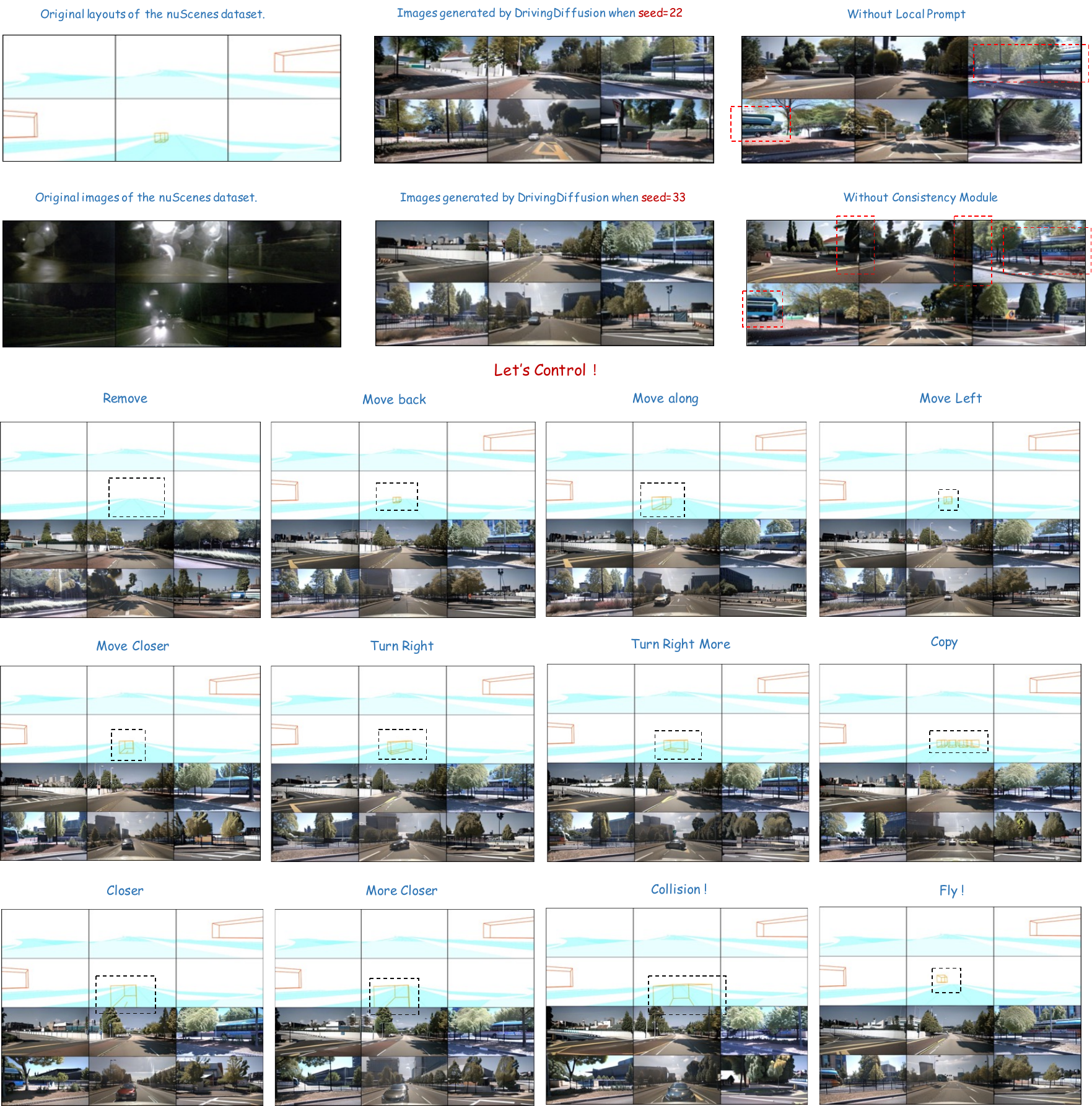}
\caption{The generation effectiveness and the ability of the layout control model under different conditions. }
\centering
\label{fig:control}
% \vspace{-10pt}
\end{figure*}
%------------------------FIG---------

\paragraph{Effectiveness of Consistency Module and Local Prompt.}
To verify the necessity of the consistency design and the local prompt, we conducted ablation experiments on the consistency module and the local prompt. We evaluated both the global quality by FID and FVD under different settings and the instance quality by Object NDS.

Compared to Setting (a) and (b), we observed a decrease in FID and FVD by 0.63 and 55, respectively. This indicates that the consistency module significantly improves the overall image and video quality, with a higher impact on FVD. This may be due to the feedback of cross-frame consistency on FVD, as it directly affects the temporal coherence of the video. On the other hand, the impact on FID is more indirect, as it mainly measures the similarity between generated and real images.

Comparing Setting (a) and (c), we found that the local prompt significantly improves the instance quality, with Object NDS increasing from 28.1 to 32.6. This demonstrates that the local prompt effectively guides the generation process to focus on specific objects, leading to better instance-level performance.

Finally, Setting (d) shows that the combination of the consistency module and the local prompt achieves the optimal generation quality: FID, FVD, and NDS reached 15.83, 332, and 33.1, respectively. By jointly considering global consistency and local guidance, our approach effectively balances the overall image/video quality and instance-level performance. This further validates the necessity of incorporating both consistency design and local prompt in our framework.

\paragraph{Visualization of Difference Settings.}

% As shown in the upper part of Figure \ref{fig:control}, we first compare the nuScenes original image with the multi-view image generated under different seed, which shows the variety of generated results. Then we shows the importance of our main design. As you can see, instances (longer cars) are less productive after the Local Prompt is removed, and there is no consistency between different perspectives after the Consistency Module is removed. In the lower part of the figure, we show the result of the original layout is disturbed. Our method can even simulate the collision scene and the car in the air.
In the upper portion of Figure \ref{fig:control}, we compare the original nuScenes image with multi-view images generated using various seeds to demonstrate the diversity of our results. Next, we highlight the significance of our core designs. As evident from the figure, the removal of the Local Prompt leads to a decrease in the quality of instances, while the elimination of the Consistency Module results in view-inconsistent perspectives. In the lower part of the figure, we present the outcome of disturbing the original layout. Our method even is capable of simulating collision scenes and cars suspended in the air.

%------------------------------------------------------------------------
\section{Conclusion}
\label{sec:conclusion}
Our work is the first to realize multi-view video generation of autonomous driving scenes and implement precise layout control, which has great significance for autonomous driving data generation and simulation. Our paradigm can be easily extended to the depth map, point cloud, occupancy, rendered assets, etc. for control of video generation. Furthermore, our approach can be trained using autoregressive techniques on a substantial number of videos(preferably those obtained through data mining).    This not only allows us to get a model that can predict future videos but also makes it an excellent pre-trained model for subsequent downstream tasks.   This may be due to the model's enhanced understanding of driving scenarios by predicting future events.

In future work, we will explore the memory-friendly end-to-end generation of video across views, which is a question of how to compress the extra dimensions.  In addition, we believe that NerF-based approaches can provide better spatial and temporal consistency, and we will explore introducing Nerf to help improve the quality of our multi-view video generation.

% Our work shows great potential in both autonomous driving data generation and simulation. Our work can expand to layout in formats such as depth chart, point cloud, occupancy, etc. to video generation. We will try to solve sim2real by using the above intermediate state as control after constructing simulation scenes.

% We will explore the memory-friendly end-to-end generation of cross-view sequential videos by conditionally diffusing the view and time dimensions after compression respectively, and finally restoring the original dimensions in the form of tensor products. In addition, we believe that Nerf-based methods can provide better spatial and temporal consistency, and we will explore the introduction of Nerf to help improve the quality of our multi-view video generation. 

% Furthermore, our findings reveal that conducting autoregressive training on a considerable amount of videos, preferably those obtained through data mining, not only allows us to obtain a model capable of predicting future videos but also enables it to serve as an excellent pre-training model for subsequent downstream tasks.  This is likely attributed to the model's enhanced comprehension of driving scenes through predicting future events. 

{\small
\bibliographystyle{ieee_fullname}
\bibliography{main}

\begin{thebibliography}{10}\itemsep=-1pt

\bibitem{blattmann2023align}
Andreas Blattmann, Robin Rombach, Huan Ling, Tim Dockhorn, Seung~Wook Kim, Sanja Fidler, and Karsten Kreis.
\newblock Align your latents: High-resolution video synthesis with latent diffusion models.
\newblock {\em arXiv preprint arXiv:2304.08818}, 2023.

\bibitem{cheng2023layoutdiffuse}
Jiaxin Cheng, Xiao Liang, Xingjian Shi, Tong He, Tianjun Xiao, and Mu Li.
\newblock Layoutdiffuse: Adapting foundational diffusion models for layout-to-image generation.
\newblock {\em arXiv preprint arXiv:2302.08908}, 2023.

\bibitem{ding2022cogview2}
Ming Ding, Wendi Zheng, Wenyi Hong, and Jie Tang.
\newblock Cogview2: Faster and better text-to-image generation via hierarchical transformers.
\newblock {\em arXiv preprint arXiv:2204.14217}, 2022.

\bibitem{fang2023tbp}
Shaoheng Fang, Zi Wang, Yiqi Zhong, Junhao Ge, Siheng Chen, and Yanfeng Wang.
\newblock Tbp-former: Learning temporal bird's-eye-view pyramid for joint perception and prediction in vision-centric autonomous driving.
\newblock {\em arXiv preprint arXiv:2303.09998}, 2023.

\bibitem{vip3d}
Junru Gu, Chenxu Hu, Tianyuan Zhang, Xuanyao Chen, Yilun Wang, Yue Wang, and Hang Zhao.
\newblock Vip3d: End-to-end visual trajectory prediction via 3d agent queries.
\newblock {\em arXiv preprint arXiv:2208.01582}, 2022.

\bibitem{he2021context}
Sen He, Wentong Liao, Michael~Ying Yang, Yongxin Yang, Yi-Zhe Song, Bodo Rosenhahn, and Tao Xiang.
\newblock Context-aware layout to image generation with enhanced object appearance.
\newblock In {\em Proceedings of the IEEE/CVF conference on computer vision and pattern recognition}, pages 15049--15058, 2021.

\bibitem{ho2020denoising}
Jonathan Ho, Ajay Jain, and Pieter Abbeel.
\newblock Denoising diffusion probabilistic models.
\newblock {\em Advances in Neural Information Processing Systems}, 33:6840--6851, 2020.

\bibitem{ho2022video}
Jonathan Ho, Tim Salimans, Alexey Gritsenko, William Chan, Mohammad Norouzi, and David~J Fleet.
\newblock Video diffusion models.
\newblock {\em arXiv:2204.03458}, 2022.

\bibitem{hong2022cogvideo}
Wenyi Hong, Ming Ding, Wendi Zheng, Xinghan Liu, and Jie Tang.
\newblock Cogvideo: Large-scale pretraining for text-to-video generation via transformers.
\newblock {\em arXiv preprint arXiv:2205.15868}, 2022.

\bibitem{hu2021fiery}
Anthony Hu, Zak Murez, Nikhil Mohan, Sof{\'\i}a Dudas, Jeffrey Hawke, Vijay Badrinarayanan, Roberto Cipolla, and Alex Kendall.
\newblock Fiery: future instance prediction in bird's-eye view from surround monocular cameras.
\newblock In {\em Proceedings of the IEEE/CVF International Conference on Computer Vision}, pages 15273--15282, 2021.

\bibitem{qd3dt}
Hou-Ning Hu, Yung-Hsu Yang, Tobias Fischer, Trevor Darrell, Fisher Yu, and Min Sun.
\newblock Monocular quasi-dense 3d object tracking.
\newblock {\em IEEE Transactions on Pattern Analysis and Machine Intelligence}, 45(2):1992--2008, 2022.

\bibitem{pip}
Bo Jiang, Shaoyu Chen, Xinggang Wang, Bencheng Liao, Tianheng Cheng, Jiajie Chen, Helong Zhou, Qian Zhang, Wenyu Liu, and Chang Huang.
\newblock Perceive, interact, predict: Learning dynamic and static clues for end-to-end motion prediction.
\newblock {\em arXiv preprint arXiv:2212.02181}, 2022.

\bibitem{lang2019pointpillars}
Alex~H Lang, Sourabh Vora, Holger Caesar, Lubing Zhou, Jiong Yang, and Oscar Beijbom.
\newblock Pointpillars: Fast encoders for object detection from point clouds.
\newblock In {\em CVPR}, pages 12697--12705, 2019.

\bibitem{li2022bevformer}
Zhiqi Li, Wenhai Wang, Hongyang Li, Enze Xie, Chonghao Sima, Tong Lu, Qiao Yu, and Jifeng Dai.
\newblock Bevformer: Learning bird's-eye-view representation from multi-camera images via spatiotemporal transformers.
\newblock {\em arXiv preprint arXiv:2203.17270}, 2022.

\bibitem{li2021image}
Zejian Li, Jingyu Wu, Immanuel Koh, Yongchuan Tang, and Lingyun Sun.
\newblock Image synthesis from layout with locality-aware mask adaption.
\newblock In {\em Proceedings of the IEEE/CVF International Conference on Computer Vision}, pages 13819--13828, 2021.

\bibitem{pnpnet}
Ming Liang, Bin Yang, Wenyuan Zeng, Yun Chen, Rui Hu, Sergio Casas, and Raquel Urtasun.
\newblock Pnpnet: End-to-end perception and prediction with tracking in the loop.
\newblock In {\em Proceedings of the IEEE/CVF Conference on Computer Vision and Pattern Recognition}, pages 11553--11562, 2020.

\bibitem{liu2022petr}
Yingfei Liu, Tiancai Wang, Xiangyu Zhang, and Jian Sun.
\newblock Petr: Position embedding transformation for multi-view 3d object detection.
\newblock {\em arXiv preprint arXiv:2203.05625}, 2022.

\bibitem{liu2022bevfusion}
Zhijian Liu, Haotian Tang, Alexander Amini, Xinyu Yang, Huizi Mao, Daniela Rus, and Song Han.
\newblock Bevfusion: Multi-task multi-sensor fusion with unified bird's-eye view representation.
\newblock {\em arXiv preprint arXiv:2205.13542}, 2022.

\bibitem{nichol2021glide}
Alex Nichol, Prafulla Dhariwal, Aditya Ramesh, Pranav Shyam, Pamela Mishkin, Bob McGrew, Ilya Sutskever, and Mark Chen.
\newblock Glide: Towards photorealistic image generation and editing with text-guided diffusion models.
\newblock {\em arXiv preprint arXiv:2112.10741}, 2021.

\bibitem{park2019semantic}
Taesung Park, Ming-Yu Liu, Ting-Chun Wang, and Jun-Yan Zhu.
\newblock Semantic image synthesis with spatially-adaptive normalization.
\newblock In {\em Proceedings of the IEEE/CVF conference on computer vision and pattern recognition}, pages 2337--2346, 2019.

\bibitem{parmar2021cleanfid}
Gaurav Parmar, Richard Zhang, and Jun-Yan Zhu.
\newblock On aliased resizing and surprising subtleties in gan evaluation.
\newblock In {\em CVPR}, 2022.

\bibitem{lss}
Jonah Philion and Sanja Fidler.
\newblock Lift, splat, shoot: Encoding images from arbitrary camera rigs by implicitly unprojecting to 3d.
\newblock In {\em ECCV}, 2020.

\bibitem{clip}
Alec Radford, Jong~Wook Kim, Chris Hallacy, Aditya Ramesh, Gabriel Goh, Sandhini Agarwal, Girish Sastry, Amanda Askell, Pamela Mishkin, Jack Clark, et~al.
\newblock Learning transferable visual models from natural language supervision.
\newblock In {\em International conference on machine learning}, pages 8748--8763. PMLR, 2021.

\bibitem{ramesh2022hierarchical}
Aditya Ramesh, Prafulla Dhariwal, Alex Nichol, Casey Chu, and Mark Chen.
\newblock Hierarchical text-conditional image generation with clip latents.
\newblock {\em arXiv preprint arXiv:2204.06125}, 2022.

\bibitem{dalle}
Aditya Ramesh, Mikhail Pavlov, Gabriel Goh, Scott Gray, Chelsea Voss, Alec Radford, Mark Chen, and Ilya Sutskever.
\newblock Zero-shot text-to-image generation.
\newblock In {\em International Conference on Machine Learning}, pages 8821--8831. PMLR, 2021.

\bibitem{rombach2022high}
Robin Rombach, Andreas Blattmann, Dominik Lorenz, Patrick Esser, and Bj{\"o}rn Ommer.
\newblock High-resolution image synthesis with latent diffusion models.
\newblock In {\em CVPR}, pages 10684--10695, 2022.

\bibitem{saharia2022photorealistic}
Chitwan Saharia, William Chan, Saurabh Saxena, Lala Li, Jay Whang, Emily Denton, Seyed Kamyar~Seyed Ghasemipour, Burcu~Karagol Ayan, S~Sara Mahdavi, Rapha~Gontijo Lopes, et~al.
\newblock Photorealistic text-to-image diffusion models with deep language understanding.
\newblock {\em arXiv preprint arXiv:2205.11487}, 2022.

\bibitem{salimans2016improved}
Tim Salimans, Ian Goodfellow, Wojciech Zaremba, Vicki Cheung, Alec Radford, and Xi Chen.
\newblock Improved techniques for training gans.
\newblock {\em Advances in neural information processing systems}, 29, 2016.

\bibitem{singer2022make}
Uriel Singer, Adam Polyak, Thomas Hayes, Xi Yin, Jie An, Songyang Zhang, Qiyuan Hu, Harry Yang, Oron Ashual, Oran Gafni, et~al.
\newblock Make-a-video: Text-to-video generation without text-video data.
\newblock {\em arXiv preprint arXiv:2209.14792}, 2022.

\bibitem{song2020denoising}
Jiaming Song, Chenlin Meng, and Stefano Ermon.
\newblock Denoising diffusion implicit models.
\newblock {\em arXiv preprint arXiv:2010.02502}, 2020.

\bibitem{swerdlow2023street}
Alexander Swerdlow, Runsheng Xu, and Bolei Zhou.
\newblock Street-view image generation from a bird's-eye view layout.
\newblock {\em arXiv preprint arXiv:2301.04634}, 2023.

\bibitem{sylvain2021object}
Tristan Sylvain, Pengchuan Zhang, Yoshua Bengio, R~Devon Hjelm, and Shikhar Sharma.
\newblock Object-centric image generation from layouts.
\newblock In {\em Proceedings of the AAAI Conference on Artificial Intelligence}, volume~35, pages 2647--2655, 2021.

\bibitem{unterthiner2018towards}
Thomas Unterthiner, Sjoerd Van~Steenkiste, Karol Kurach, Raphael Marinier, Marcin Michalski, and Sylvain Gelly.
\newblock Towards accurate generative models of video: A new metric \& challenges.
\newblock {\em arXiv preprint arXiv:1812.01717}, 2018.

\bibitem{van2017neural}
Aaron Van Den~Oord, Oriol Vinyals, et~al.
\newblock Neural discrete representation learning.
\newblock {\em Advances in neural information processing systems}, 30, 2017.

\bibitem{2309.09777}
Xiaofeng Wang, Zheng Zhu, Guan Huang, Xinze Chen, and Jiwen Lu.
\newblock Drivedreamer: Towards real-world-driven world models for autonomous driving, 2023.

\bibitem{wu2021godiva}
Chenfei Wu, Lun Huang, Qianxi Zhang, Binyang Li, Lei Ji, Fan Yang, Guillermo Sapiro, and Nan Duan.
\newblock Godiva: Generating open-domain videos from natural descriptions.
\newblock {\em arXiv preprint arXiv:2104.14806}, 2021.

\bibitem{wu2022nuwa}
Chenfei Wu, Jian Liang, Lei Ji, Fan Yang, Yuejian Fang, Daxin Jiang, and Nan Duan.
\newblock N{\"u}wa: Visual synthesis pre-training for neural visual world creation.
\newblock In {\em Computer Vision--ECCV 2022: 17th European Conference, Tel Aviv, Israel, October 23--27, 2022, Proceedings, Part XVI}, pages 720--736. Springer, 2022.

\bibitem{xiong2023cape}
Kaixin Xiong, Shi Gong, Xiaoqing Ye, Xiao Tan, Ji Wan, Errui Ding, Jingdong Wang, and Xiang Bai.
\newblock Cape: Camera view position embedding for multi-view 3d object detection.
\newblock {\em arXiv preprint arXiv:2303.10209}, 2023.

\bibitem{yu2021vector}
Jiahui Yu, Xin Li, Jing~Yu Koh, Han Zhang, Ruoming Pang, James Qin, Alexander Ku, Yuanzhong Xu, Jason Baldridge, and Yonghui Wu.
\newblock Vector-quantized image modeling with improved vqgan.
\newblock {\em arXiv preprint arXiv:2110.04627}, 2021.

\bibitem{yu2022scaling}
Jiahui Yu, Yuanzhong Xu, Jing~Yu Koh, Thang Luong, Gunjan Baid, Zirui Wang, Vijay Vasudevan, Alexander Ku, Yinfei Yang, Burcu~Karagol Ayan, et~al.
\newblock Scaling autoregressive models for content-rich text-to-image generation.
\newblock {\em arXiv preprint arXiv:2206.10789}, 2022.

\bibitem{zhang2023adding}
Lvmin Zhang and Maneesh Agrawala.
\newblock Adding conditional control to text-to-image diffusion models.
\newblock {\em arXiv preprint arXiv:2302.05543}, 2023.

\bibitem{mutr3d}
Tianyuan Zhang, Xuanyao Chen, Yue Wang, Yilun Wang, and Hang Zhao.
\newblock Mutr3d: A multi-camera tracking framework via 3d-to-2d queries.
\newblock In {\em Proceedings of the IEEE/CVF Conference on Computer Vision and Pattern Recognition}, pages 4537--4546, 2022.

\bibitem{bytetrackv2}
Yifu Zhang, Xinggang Wang, Xiaoqing Ye, Wei Zhang, Jincheng Lu, Xiao Tan, Errui Ding, Peize Sun, and Jingdong Wang.
\newblock Bytetrackv2: 2d and 3d multi-object tracking by associating every detection box.
\newblock {\em arXiv preprint arXiv:2303.15334}, 2023.

\bibitem{zheng2023layoutdiffusion}
Guangcong Zheng, Xianpan Zhou, Xuewei Li, Zhongang Qi, Ying Shan, and Xi Li.
\newblock Layoutdiffusion: Controllable diffusion model for layout-to-image generation.
\newblock {\em arXiv preprint arXiv:2303.17189}, 2023.

\bibitem{zhou2022cross}
Brady Zhou and Philipp Kr{\"a}henb{\"u}hl.
\newblock Cross-view transformers for real-time map-view semantic segmentation.
\newblock In {\em Proceedings of the IEEE/CVF Conference on Computer Vision and Pattern Recognition}, pages 13760--13769, 2022.

\end{thebibliography}
}
\end{document}